\documentclass[runningheads]{llncs}
\usepackage{graphicx}
\usepackage{comment}
\usepackage{amsmath,amssymb} % define this before the line numbering.
\DeclareMathOperator*{\argmin}{argmin}
\usepackage{color}
\usepackage{epsfig}
\usepackage{subfigure}
\usepackage{mathtools}
\usepackage{booktabs}
\usepackage[ruled,vlined]{algorithm2e}
\SetArgSty{textnormal}
\usepackage[pagebackref=true,breaklinks=true,letterpaper=true,colorlinks,bookmarks=false]{hyperref}

\begin{document}
% \linenumbers
\pagestyle{headings}
\mainmatter
\def\ECCVSubNumber{4072}  % Insert your submission number here

\title{Inclusive GAN: Improving Data and Minority Coverage in Generative Models} % Replace with your title

\begin{comment}
\titlerunning{ECCV-20 submission ID \ECCVSubNumber} 
\authorrunning{ECCV-20 submission ID \ECCVSubNumber} 
\author{Anonymous ECCV submission}
\institute{Paper ID \ECCVSubNumber}
\end{comment}
%******************

%\begin{comment}
\titlerunning{Inclusive GAN}
\author{Ning Yu\inst{1,2} \and
Ke Li\inst{3,5,6} \and
Peng Zhou\inst{1} \\
Jitendra Malik\inst{3} \and
Larry Davis\inst{1} \and
Mario Fritz\inst{4}}
\authorrunning{Ning Yu et al.}
\institute{University of Maryland, College Park, United States \and
Max Planck Institute for Informatics, Saarbr\"{u}cken, Germany \and
University of California, Berkeley, United States \and
CISPA Helmholtz Center for Information Security, Saarbr\"{u}cken, Germany  \and
Institute for Advanced Study, Princeton, United States \and
Google, Seattle, United States\\
\email{ningyu@mpi-inf.mpg.de} \hspace{3ex} \email{ke.li@eecs.berkeley.edu} \hspace{3ex} \email{pengzhou@cs.umd.edu} \\
\email{malik@eecs.berkeley.edu} \hspace{3ex} \email{lsd@cs.umd.edu} \hspace{3ex} \email{fritz@cispa.saarland}}
%\end{comment}
%******************
\maketitle

\begin{abstract}
Generative Adversarial Networks (GANs) have brought about rapid progress towards generating photorealistic images. Yet the equitable allocation of their modeling capacity among subgroups has received less attention, which could lead to potential biases against underrepresented minorities if left uncontrolled. In this work, we first formalize the problem of minority inclusion as one of data coverage, and then propose to improve data coverage by harmonizing adversarial training with reconstructive generation. The experiments show that our method outperforms the existing state-of-the-art methods in terms of data coverage on both seen and unseen data. We develop an extension that allows explicit control over the minority subgroups that the model should ensure to include, and validate its effectiveness at little compromise from the overall performance on the entire dataset. Code, models, and supplemental videos are available at \href{https://github.com/ningyu1991/InclusiveGAN.git}{GitHub}.

\end{abstract}

\section{Introduction}

Photorealistic image generation has increasingly become reality, thanks to the emergence of large-scale datasets~\cite{deng2009imagenet,liu2015deep,yu15lsun} and deep generative models~\cite{kingma2013auto,goodfellow2014generative,larsen2015autoencoding,li2018implicit}. However, these advances have come at a cost: there could be potential biases in the learned model against underrepresented data subgroups~\cite{xu2018fairgan,zhao2018bias,sattigeri2019fairness,Grover2019FairGM,grover2019bias}. The biases are rooted in the inevitable imbalance in the dataset~\cite{ryu2017inclusivefacenet}, which are preserved or even exacerbated by the generative models~\cite{zhao2018bias}. In particular, reconstructive (non-adversarial) generative models like variational autoencoders (VAEs)~\cite{kingma2013auto,rezende2014stochastic} can preserve data biases against minorities due to their objective of reproducing the frequencies images occur in the dataset, while adversarial generative models (GANs)~\cite{goodfellow2014generative,radford2015unsupervised,dumoulin2016adversarially,donahue2016adversarial,miyato2018spectral,karras2017progressive,brock2018large,karras2019style,karras2020analyzing} can implicitly disregard infrequent images due to the well-established problem of mode collapse~\cite{srivastava2017veegan,li2018implicit}, thereby further introducing model biases on top of data biases. This issue is particularly acute from the perspective of minority inclusion, because training data associated with minority subgroups by definition do not form dominant modes. Consequently, data from minority groups are rare to begin with, and would not be capable of being produced by the generative model at all due to mode collapse.

\begin{figure*}[!t]
\centering
\includegraphics[width=\textwidth]{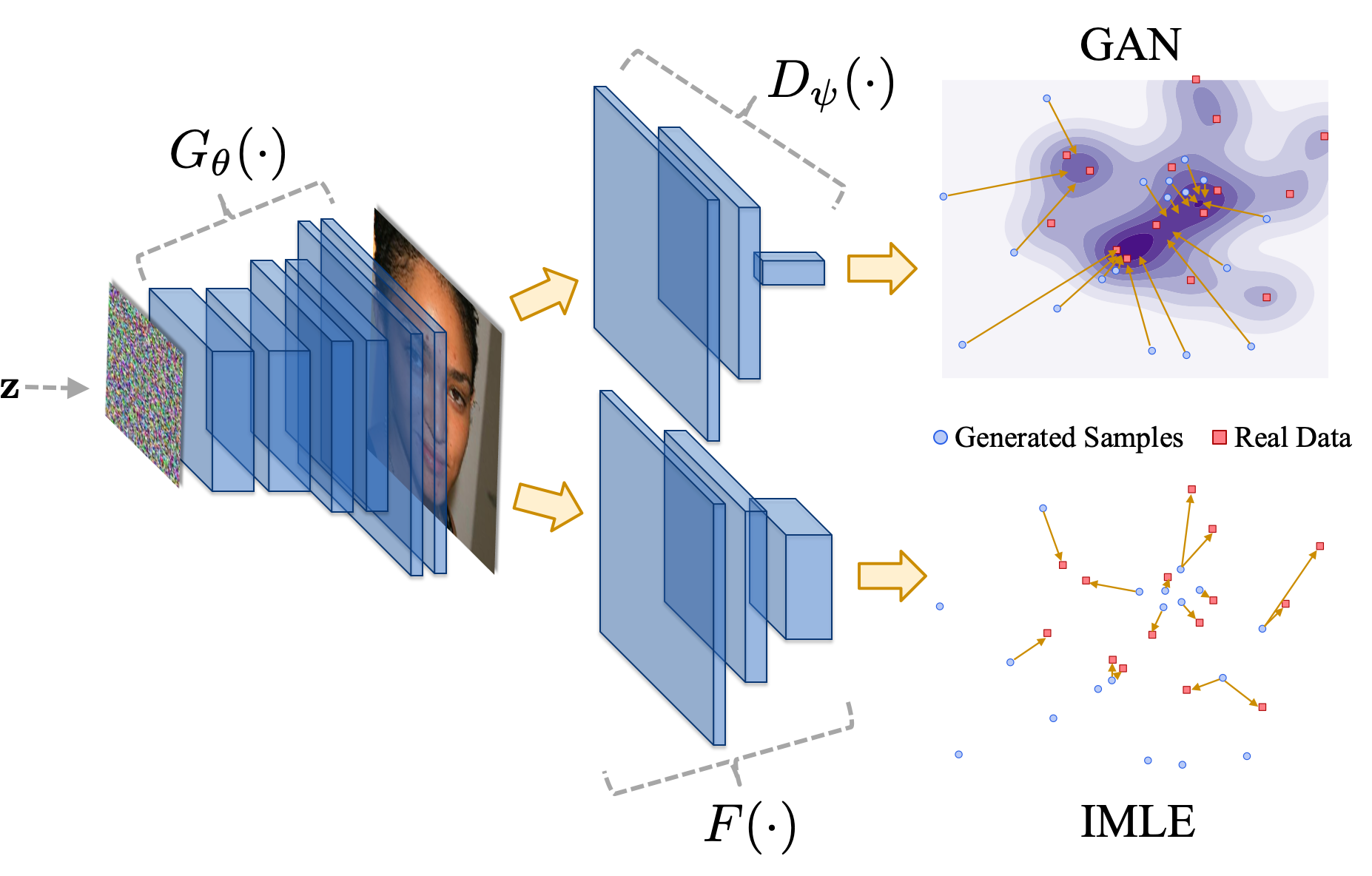}
\caption{The diagram of our method. It harmonizes adversarial (GAN) and reconstructive (IMLE) training in one framework without introducing an auxiliary encoder. GAN guides arbitrary sampling towards generating realistic appearances approximate to some real data while IMLE ensures data coverage where there are always generated samples approximate to each real data. See Section~\ref{sec:imle_gan} for more details where $G_\theta$ and $D_\psi$ represent the trainable generator and discriminator in a GAN, and $F$ represents a distant metric, in some cases, a pre-trained neural network.}
\label{fig:teaser}
\end{figure*}

In this work, we aim to improve the \emph{comprehensive} performance of the state-of-the-art generative models, with a specific focus on their coverage of minority subgroups. We start with an empirical study on the correlation between data biases and model biases, and then formalize the objective of alleviating model bias in terms of improving data coverage, in particular over the minority subgroups. We propose a new method known as IMLE-GAN that achieves competitive image quality while ensuring improved coverage of minority groups.  

Our method harmonizes adversarial and reconstructive generative models, in the process combining the benefits of both. Adversarial models have evolved to generate photorealistic results, whereas reconstructive models offer guarantees on data coverage. We build upon one of the state-of-the-art implementations of adversarial models, i.e., StyleGAN2~\cite{karras2020analyzing}, and incorporate it with the Implicit Maximum Likelihood Estimation (IMLE) framework~\cite{li2018implicit}, which is at its core reconstructive. See Figure~\ref{fig:teaser} for a diagram. 

Different from the existing hybrid generative models~\cite{larsen2015autoencoding,srivastava2017veegan,rosca2017variational,bhattacharyya2019best} that require training an auxiliary encoder network alongside a vanilla GAN, our method operates purely with the standard components of a GAN. This brings two main benefits: (1) it sidesteps the complication from combining the minimax objective used by adversarial models and the pure minimization objective used by reconstructive models, and (2) it avoids carrying over the practical issues of training auxiliary encoder, like posterior collapse~\cite{bowman2016generating,kim2018semi}, which can cause the regression-to-the-mean problem, leading to blurry images.

We validate our method with thorough experiments and demonstrate more comprehensive data coverage that goes beyond that of existing state-of-the-art methods. In addition, our method can be flexibly adapted to ensure the inclusion of specified minority subgroups, which cannot be easily achieved in the context of existing methods. 

\noindent\textbf{Contributions.} We summarize our main contributions as follows: (1) we study the problem of underrepresented minority inclusion and formalize it as a data coverage problem in generative modeling; (2) we present a novel paradigm of harmonizing adversarial and reconstructive modeling for improving data coverage; (3) our experiments set up a new suite of state-of-the-art performance in terms of covering both seen and unseen data; and (4) we develop an effective extension of our technique to ensure inclusion of the specified minority subgroups.

\section{Related Work}
\label{sec:related_work}

\noindent\textbf{Bias mitigation efforts for machine learning.} Bias in machine learning results from data imbalance, which can be detected and alleviated by three categories of approaches: The pre-process approaches that purify data from bias before training~\cite{calders2009building,dwork2012fairness,feldman2015certifying,zhang2016causal}, the in-process approaches that enforce fairness during training with constraints or regularization in the objectives~\cite{kamishima2011fairness,zafar2015fairness,ryu2017inclusivefacenet,zhao2017men}, and the post-process approaches that adjust the output from a learned model~\cite{kamiran2010discrimination,hardt2016equality}. A comprehensive survey~\cite{mehrabi2019survey} articulates this taxonomy. These approaches target biases in classification and cannot be adapted to generative modeling.

\noindent\textbf{Bias mitigation efforts for generative models.} There have been relatively few papers~\cite{xu2018fairgan,zhao2018bias,sattigeri2019fairness,Grover2019FairGM,grover2019bias} that focus on biases in generative models. \cite{xu2018fairgan,sattigeri2019fairness,Grover2019FairGM}, motivated from benefiting a downstream classifier, mainly aim for fair generation conditioned on attribute inputs, in terms of yielding allocative decisions and/or removing the correlation between generation and attribute conditions. \cite{zhao2018bias} focuses on understanding the inductive bias so as to investigate the generalization of generative models. \cite{grover2019bias} proposes an importance weighting strategy to compensate for the biases of learned generative models. Different from their goals and solutions that equalize performance across different data subgroups possibly at the cost of overall performance, we instead aim to improve the overall data coverage, with a specific purpose of ensuring more significant gains over the underrepresented minorities.

\noindent\textbf{Data coverage in GANs.} GANs are finicky to train because of the minimax formulation and the alternating gradient ascent-descent. In addition, GANs are known to exhibit mode collapse, where the generator only learns to generate a subset of the modes of the underlying data distribution. To alleviate mode collapse in GANs, some methods propose to improve the minimax loss function~\cite{metz2016unrolled,arjovsky2017wasserstein,gulrajani2017improved,adler2018banach}, some methods apply constraints or regularization terms along with the minimax objectives~\cite{chen2016infogan,berthelot2017began,tran2018dist,dsganICLR2019,lin2018pacgan}, and some other methods aim to modify the discriminator designs~\cite{warde2016improving,zhao2016energy,miyato2018spectral,peng2019variational}. These directions are orthogonal to our research while, in principle, demonstrate less effective data coverage than the hybrid models below.

\noindent\textbf{Data coverage in hybrid generative models.} Reconstructive (non-adversarial) generative models like variational autoencoders (VAEs)~\cite{kingma2013auto,rezende2014stochastic}, on the other hand, are more successful at data coverage because they explicitly try to maximize a lower bound on the likelihood of the real data. This motivates a variety of designs for hybrid models that combine reconstruction and adversarial training. $\alpha$-GAN~\cite{rosca2017variational} is trained to reconstruct pixels while VAEGAN~\cite{larsen2015autoencoding} is trained to reconstruct discriminator features. ALI~\cite{dumoulin2016adversarially}, BiGAN~\cite{donahue2016adversarial}, and SVAE~\cite{chen2018symmetric} propose to instead jointly match the bidirectional mappings between data and latent distributions. VEEGAN~\cite{srivastava2017veegan} is designed with reconstruction in the latent space, in the purpose of avoiding the metric dilemma in the data space. Hybrid models benefit for mode coverage, but deteriorate generation fidelity in practice, because of their dependency on auxiliary encoder networks. In contrast, our method follows the idea of hybrid models, but avoids an encoder network and instead apply all training back-propagation through the generator. A recent non-adversarial generative framework, Implicit Maximum Likelihood Estimation (IMLE)~\cite{li2018implicit}, satisfies our design. We discuss more about the advantages of IMLE in Section~\ref{sec:imle}.

\section{Inclusive GAN for Data and Minority Coverage}

Our method is a novel paradigm of harmonizing the strengths of adversarial (Section~\ref{sec:gans}) and reconstructive generative models (Section~\ref{sec:imle}) that avoids mode collapse. The harmonization efforts (Section~\ref{sec:imle_gan}) are necessary and non-trivial due to the incompatibility between the two, which is validated in the supplementary material. In Section~\ref{sec:imle_gan_minority} we show the straightforward adaptation of our method to improve minority inclusion.

\subsection{Adversarial Generation: GANs}
\label{sec:gans}

Photorealistic image generation can be viewed as the problem of sampling from the unknown probability distribution of real-world images. Generative Adversarial Networks (GANs)~\cite{goodfellow2014generative} introduce an elegant solution for distribution estimation, which is formulated as a discriminative classification problem, and enables supervised learning methods to be used for this task.

A GAN consists of two deep neural networks: a generator $G_\theta: \mathbb{R}^d \mapsto \mathbb{R}^D$ and a discriminator $D_\psi: \mathbb{R}^D \mapsto [0, 1]$. The generator maps a latent noise vector $\mathbf{z} \sim \mathcal{N}(\mathbf{0},\mathbf{I}_d)$ to an image, and the discriminator predicts the probability that the image it sees is real. The real ground truth images are denoted as $\mathbf{x}\sim \hat{p}(\mathbf{x})$, sampled from an unknown distribution $\hat{p}(\mathbf{x})$. The discriminator is trained to maximize classification accuracy while the generator is trained to produce images that can fool the discriminator. More precisely, the objective is shown in Eq.~\ref{eq:adv}:
\begin{equation}
\min_\theta\max_\psi L^{\textit{adv}}(\theta,\psi) = \mathbb{E}_{\mathbf{x}\sim \hat{p}(\mathbf{x})}\left[\log D_\psi(\mathbf{x})\right] + \mathbb{E}_{\mathbf{z}\sim \mathcal{N}(\mathbf{0},\mathbf{I}_d)}\left[\log(1-D_\psi(G_\theta(\mathbf{z})))\right]
\label{eq:adv}
\end{equation}

Unfortunately, GANs are unstable to train and suffer from mode collapse: While each generated sample gets to pick a mode it is drawn to, each mode does not get to pick a generated sample. After training, the generator will not be able to generate samples around the ``unpopular'' modes. 

Minority modes are precisely the ``unpopular'' modes that are more likely to be collapsed. As shown in Section~\ref{sec:bias_study} and Figure~\ref{fig:counts_std_dists}, minority subgroups with diverse appearances indeed bring more challenges to generative modeling and are allocated worse coverage compared to the others. Therefore, we propose to leverage reconstructive models to improve the coverage of minority subgroups.

\subsection{Reconstructive Generation: IMLE}
\label{sec:imle}

Our novel paradigm is based on a recent reconstructive framework, Implicit Maximum Likelihood Estimation (IMLE)~\cite{li2018implicit}, that favors complete mode coverage. IMLE avoids mode collapse by reversing the direction in which generated samples are matched to real modes. In GANs, each generated sample is effectively matched to a real mode. In IMLE, each real mode is matched to a generated sample. This ensures that all real modes, including each underrepresented minority mode, are matched, and no real mode is left out.

Mathematically, IMLE tackles the optimization problem in Eq.~\ref{eq:imle}:
\begin{align}
& \min_\theta \mathbb{E}_{\mathbf{z}_1,\ldots,\mathbf{z}_{m}\sim \mathcal{N}(\mathbf{0},\mathbf{I}_d)}\left[\mathbb{E}_{\mathbf{x}\sim \hat{p}(\mathbf{x})}\left[\min_{i\in\{1,\ldots,m\}}\Vert G_\theta(\mathbf{z}_i)-\mathbf{x}\Vert_2^2\right]\right] \label{eq:imle} \\
= & \min_\theta \mathbb{E}_{\mathbf{z}_1,\ldots,\mathbf{z}_{m}\sim \mathcal{N}(\mathbf{0},\mathbf{I}_d)}\left[\mathbb{E}_{\mathbf{x}\sim \hat{p}(\mathbf{x})}\left[\Vert G_\theta(\mathbf{z}^*(\mathbf{x}))-\mathbf{x}\Vert_2^2\right]\right]\mbox{,} \label{eq:imle2} \\
\mbox{where }\mathbf{z}^*= & \argmin_{i\in\{1,\ldots,m\}}\Vert G_\theta(\mathbf{z}_i) - \mathbf{x}\Vert_2^2
\end{align}

The joint optimization is achieved by alternating between the two decoupled phases until convergence. The first phase corresponds to the inner optimization, where we search for each $\mathbf{x}$ the optimal $\mathbf{z}^*(\mathbf{x})$ from the latent vector candidates, given a fixed $G_\theta$. This is implemented by the Prioritized DCI~\cite{li2017fast}, a fast nearest neighbor search algorithm. The second phase corresponds to the outer optimization, where we train the generator in the regular back-propagation manner, given pairs of $(\mathbf{x}, \mathbf{z}^*(\mathbf{x}))$.

One significant advantage of IMLE over the other reconstructive models is the elimination of the need for an auxiliary encoder. The encoder encourages mode coverage but at the cost of either deviating the latent sampling distribution from the original prior (in VAEGAN~\cite{larsen2015autoencoding}) or absorbing the training gradients before substantially back-propagating to the generator (in VEEGAN~\cite{srivastava2017veegan}). Unlike them, IMLE directly samples latent vector from a natural prior during training and encourages explicit reconstruction fully upon the generator.

\subsection{Harmonizing Adversarial and Reconstructive Generation: IMLE-GAN}
\label{sec:imle_gan}

Below we propose a way to harmonize adversarial training with the IMLE framework, so as to ensure both generation quality (precision) and coverage (recall) simultaneously.

The vanilla hybrid model between IMLE and GAN is to directly add the adversarial loss in Eq.~\ref{eq:adv} to the non-adversarial loss in Eq.~\ref{eq:imle}. This has two problems because of (1) differences in the domains over which latent vectors are sampled and (2) differences in the metric spaces on which GAN and IMLE operate. For (1), in the case of GAN, a different latent vector is randomly sampled every iteration, whereas in the case of IMLE, many latent vectors are sampled at once (over which matching is performed) and are kept fixed for many iterations. The former gives up control over which data point each latent vector is asked to generate by the discriminator, but can avoid overfitting to any one latent vector. The latter explicitly controls which latent vectors are matched to data points, but can overfit to the set of matched latent vectors until they are resampled. For (2), in the case of GAN, the discriminator takes the inner product between the features and the weight vector of the last layer to produce a realism score, and so it effectively operates on features of images; on the other hand, in the case of IMLE, matching is performed on raw pixels. 

To bridge the gap in losses, we propose two adaptations that better harmonize the GAN and IMLE objectives. First, to make the domain over which latent vectors are sampled denser, we augment the matched latent vectors with random linear interpolations. Second, to make the spaces on which the two losses are computed more comparable, we measure the reconstruction loss in a deep feature space instead of pixel space, such that it contains a comparable amount and level of semantic information to that used by the discriminator. Mathematically, our goal is to optimize Eq.~\ref{eq:imle_gam}:
\begin{equation}
\min_\theta \max_\psi L^{\textit{adv}}(\theta,\psi) + \mathbb{E}_{\mathbf{z}_1,\ldots,\mathbf{z}_m\sim \mathcal{N}(\mathbf{0},\mathbf{I}_d)}\left[\lambda L^{\textit{rec}}(\theta) + \beta L^{\textit{itp}}(\theta)\right]
\label{eq:imle_gam}
\end{equation}
Here $L^{\textit{adv}}(\theta,\psi)$ is as defined in Eq.~\ref{eq:adv}, 
\begin{align}
L^{\textit{rec}}(\theta) = & \mathbb{E}_{\mathbf{x}\sim \hat{p}(\mathbf{x})}\left[\Vert F( G_\theta(\mathbf{z}^*(\mathbf{x}))) - F(\mathbf{x})\Vert_2^2\right]
\label{eq:rec} \\
\mbox{where } \mathbf{z}^*(\mathbf{x}) = & \argmin_{i\in\{1,\ldots,m\}}\Vert F(G_\theta(\mathbf{z}_i)) - F(\mathbf{x})\Vert_2^2\mbox{, } \label{eq:latent_matching} \\
\mbox{ and }L^{\textit{itp}}(\theta) = &  \mathbb{E}_{\mathbf{x},\widetilde{\mathbf{x}}\sim \hat{p}(\mathbf{x}),\alpha\sim U[0,1]}\left[\alpha \Vert F(G_\theta(\mathbf{z}^*(\alpha, \mathbf{x},\widetilde{\mathbf{x}}))) - F(\mathbf{x})\Vert_2^2 +\right. \label{eq:itp}\\ & \left.(1-\alpha) \Vert F(G_\theta(\mathbf{z}^*(\alpha, \mathbf{x},\widetilde{\mathbf{x}}))) - F(\widetilde{\mathbf{x}}) \Vert_2^2\right]
\label{eq:itp2} \\
\mbox{where } \mathbf{z}^*(\alpha, \mathbf{x},\widetilde{\mathbf{x}}) = & \alpha \mathbf{z}^*(\mathbf{x}) + (1-\alpha) \mathbf{z}^*(\widetilde{\mathbf{x}})
\label{eq:latent_itp}
\end{align}
%\begin{equation}
%\mathbf{z}^*(\alpha, \mathbf{x},\widetilde{\mathbf{x}}) = \alpha \mathbf{z}^*(\mathbf{x}) + (1-\alpha) \mathbf{z}^*(\widetilde{\mathbf{x}})
%\label{eq:latent_itp}
%\end{equation}
Here Eq.~\ref{eq:rec} generalizes Eq.~\ref{eq:imle2} by computing distance in feature space, where $F(\cdot)$ is a fixed function to compute features of images.
% and Eq.~\ref{eq:latent_matching} corresponds to the inner optimization in Eq.~\ref{eq:imle}
Eq.~\ref{eq:itp} and \ref{eq:itp2} defines the interpolation loss, which linearly interpolates between two matched latent vectors $\mathbf{z}^*(\mathbf{x}),\mathbf{z}^*(\widetilde{\mathbf{x}})$ (as shown in Eq.~\ref{eq:latent_itp}) and tries to make the image generated from the interpolated latent vector $\mathbf{z}^*(\alpha, \mathbf{x},\widetilde{\mathbf{x}})$ similar to the two ground truth images $\mathbf{x},\widetilde{\mathbf{x}}$ that correspond to the latent vectors at the endpoints. The weight on the distance to each ground truth image depends on how close the interpolated latent vector is to the endpoint, which is denoted by $\alpha$. $\lambda$ and $\beta$ are used to balance each loss term. We experiment with four possible feature spaces: raw pixels, discriminator features~\cite{larsen2015autoencoding}, Inception features~\cite{salimans2016improved}, and LPIPS features (i.e.: features such that the $\ell_2$ distance between them is equivalent to the LPIPS perceptual metric~\cite{zhang2018unreasonable}), and compare them in the the supplementary material.

\begin{algorithm}[!t]
\caption{IMLE-GAN with Minority Inclusion}
\SetAlgoLined
\KwData{Real training data $\hat{p}(\mathbf{x})$ and a specified minority subgroup $\hat{q}(\mathbf{y})$}
\KwResult{A generator $G_\theta$ with specified minority inclusion performance}
 \For{\text{epoch} = $\{1,\dots,E\}$}{
  \If{epoch \% $S == 0$}{
   Sample $\mathbf{z}_1,\ldots,\mathbf{z}_m\sim \mathcal{N}(0,\mathbf{I}_d)$ i.i.d.\;
   \For{$\mathbf{y}_j\sim\hat{q}(\mathbf{y})$}{
    $\mathbf{z}^*(\mathbf{y}_j)\gets\arg\min_{i\in\{1,\ldots,m\}}||F(G_\theta(\mathbf{z}_i))-F(\mathbf{y}_j)||_2^2$\;
   }
  }
  \For{$\mathbf{x}_k\sim\hat{p}(\mathbf{x})$ and $\mathbf{y}_i,\mathbf{y}_j\sim\hat{q}(\mathbf{y})$}{
   Sample $\mathbf{z}\sim \mathcal{N}(0,\mathbf{I}_d)$\;
   $L^{\textit{adv}}\gets \log D_\psi(\mathbf{x}_k) + \log(1-D_\psi(G_\theta(\mathbf{z})))$\;
   Sample $\mathbf{\delta}_i,\mathbf{\delta}_j\sim \mathcal{N}(0,\sigma\mathbf{I}_d)$ i.i.d.\;
   $\mathbf{z}_i^*\gets \mathbf{z}^*(\mathbf{y}_i) + \mathbf{\delta}_i$\;
   $\mathbf{z}_j^* \gets \mathbf{z}^*(\mathbf{y}_j) + \mathbf{\delta}_j$\;
   $L^{\textit{rec}}\gets \frac{1}{2}(||F(G_\theta(\mathbf{z}_i^*))- F(\mathbf{y}_i)||_2^2 + ||F(G_\theta(\mathbf{z}_j^*))-F(\mathbf{y}_j)||_2^2)$\;
   Sample $\alpha\sim U[0,1]$\;
   $\mathbf{z}_{ij}^* = \alpha\mathbf{z}_i^* + (1-\alpha)\mathbf{z}_j^*$\;
   $L^{\textit{itp}}\gets \alpha ||F(G_\theta(\mathbf{z}_{ij}^*))-F(\mathbf{y}_i)||_2^2 + (1-\alpha)||F((G_\theta(\mathbf{z}_{ij}^*))-F(\mathbf{y}_j)||_2^2$\;
   $L\gets L^{\textit{adv}} + \lambda L^{\textit{rec}} + \beta L^{\textit{itp}}$\;
   $\psi = \psi + \eta\nabla_\psi L$\;
   $\theta = \theta - \eta\nabla_\theta L$\;
  }
 }
\label{alg:imle_gan}
\end{algorithm}

\subsection{Minority Coverage in IMLE-GAN}
\label{sec:imle_gan_minority}

IMLE-GAN framework is designed to improve the overall mode coverage. One benefit compared to other hybrid models is that it is straightforward to adapt it for minority inclusion. We simply need to replace the empirical distribution over the entire dataset $\hat{p}(\mathbf{x})$ with a distribution $\hat{q}(\mathbf{x})$ whose support only covers a specified minority subgroup (i.e.: $\mathrm{supp}(\hat{q}) \subset \mathrm{supp}(\hat{p})$) in Eq.~\ref{eq:rec} and \ref{eq:itp} (for reconstructive training) and leave Eq.~\ref{eq:adv} unchanged (for adversarial training). This ensures an explicit coverage over the minority while still carrying out the approximation to the entire real data. This comes with another advantage: because $\hat{q}(\mathbf{x})$ in practice has support over a much smaller set than $\hat{p}(\mathbf{x})$, there is less data imbalance and variance within the support of $\hat{q}(\mathbf{x})$ than in $\hat{p}(\mathbf{x})$, thereby requiring less model capacity to model. As a result, covering $\hat{q}(\mathbf{x})$ should be easier than covering $\hat{p}(\mathbf{x})$, and so the perceptual quality of samples tend to improve. 

We summarize our IMLE-GAN algorithm with minority inclusion in Algorithm~\ref{alg:imle_gan}, where $E$ is the number of training epochs, $S$ indicates how often (in epochs) to update latent matching, $m$ is the pool size of the latent vector candidates, $\mathbf{\delta}_i$, $\mathbf{\delta}_j$ are the additive Gaussian perturbations, and $\eta$ is the learning rate. We provide the hyperparameter settings in the supplementary material.

\section{Experiments}

We articulate the experimental setup in Section~\ref{sec:setup}. In Section~\ref{sec:preliminary_study_mnist} we start with preliminary validation on Stacked MNIST dataset~\cite{metz2016unrolled}, an easy and interpretable task. In Section~\ref{sec:bias_study} we conduct empirical study to analyze the correlation between data bias and model bias. We then move on to the validation of our two harmonization strategies in the supplementary material. In Section~\ref{sec:comparisons_celeba} we perform comprehensive evaluation and comparisons on CelebA dataset~\cite{liu2015deep}, and finally specify minority inclusion applications in Section~\ref{sec:exp_minority_inclusion}.

\subsection{Setup}
\label{sec:setup}

\noindent\textbf{Datasets.} For preliminary study, we employ Stacked MNIST dataset~\cite{metz2016unrolled} for explicit data coverage evaluation. 240,000 RGB images in the size of 32$\times$32 are synthesized by stacking three random digit images from MNIST~\cite{lecun1998gradient} along the color channel, resulting in 1,000 explicit modes in a uniform distribution.

We conduct our main experiments on CelebA human face dataset~\cite{liu2015deep}, where the 40 binary facial attributes are used to specify minority subgroups. We sample the first 30,000 images in the size of 128$\times$128 for GAN training, and sample the last 3,000 or 30,000 images for validation.

\noindent\textbf{GAN backbone.} We build our IMLE-GAN framework on the state-of-the-art StyleGAN2~\cite{karras2020analyzing} architecture for unconditional image generation. We reuse all their default settings.

\noindent\textbf{Baseline methods.} Besides the backbone StyleGAN2~\cite{karras2020analyzing}, we also compare our method to eight techniques that show improvement in data coverage and/or generation diversity: SNGAN~\cite{miyato2018spectral}, Dist-GAN~\cite{tran2018dist}, DSGAN~\cite{dsganICLR2019}, PacGAN~\cite{lin2018pacgan}, ALI~\cite{dumoulin2016adversarially}, VAEGAN~\cite{larsen2015autoencoding}, $\alpha$-GAN~\cite{rosca2017variational}, and VEEGAN~\cite{srivastava2017veegan}. For VAEGAN which originally involves image reconstruction in the discriminator feature space, we also experiment with three other distance metrics as discussed in Section~\ref{sec:imle_gan}. For fair comparisons, we replace the original architectures used in all methods with StyleGAN2. See supplementary material for their parameter settings.

\noindent\textbf{Evaluation.} For Stacked MNIST, following~\cite{metz2016unrolled,srivastava2017veegan}, we report the number of generated modes that is detected by a pre-trained mode classifier, as well as the KL divergence between the generated mode distribution and the uniform distribution. The statistics are calculated from 240,000 randomly generated samples.

For CelebA, Fr\'{e}chet Inception Distance (FID)~\cite{heusel2017gans} is used to reflect both data quality (precision) and coverage (recall) in an entangled manner. We also explicitly measure the Precision and Recall~\cite{sajjadi2018assessing} of a generative model w.r.t. the real dataset in the Inception space. Moreover, to emphasize on instance-level data coverage, we further include Inference via Optimization Measure (IvOM)~\cite{metz2016unrolled} into our metric suite, which measures the mean retrieval error from a generative model given each query image. We also report the standard deviation of IvOM across 40 CelebA attributes, in order to evaluate the balance of generative coverage. For the generalization purpose, we evaluate over both the training set and a validation set (unseen during training). More details of the evaluation implementation are in the supplementary material.

\begin{table}[!t]
\center
\caption{Comparisons on Stacked MNIST dataset. The statistics are calculated from 240,000 randomly generated samples. We indicate for each metric whether a higher ($\Uparrow$) or lower ($\Downarrow$) value is more desirable. We highlight the best performance in \textbf{bold}.}
\begin{tabular*}{0.8\textwidth}{@{\extracolsep{\fill}} lcc}
\toprule
& \# modes (max 1000) ($\Uparrow$) & KL to uniform ($\Downarrow$)  \tabularnewline
\midrule
StyleGAN2~\cite{karras2020analyzing} & 940 & 0.424 \tabularnewline
SNGAN~\cite{miyato2018spectral} & 571 & 1.382 \tabularnewline
DSGAN~\cite{dsganICLR2019} & 955 & 0.343 \tabularnewline
PacGAN~\cite{lin2018pacgan} & 908 & 0.638 \tabularnewline
ALI~\cite{dumoulin2016adversarially} & 956 & 0.680 \tabularnewline
VAEGAN~\cite{larsen2015autoencoding} & 929 & 0.534 \tabularnewline
VEEGAN~\cite{srivastava2017veegan} & 987 & 0.310\tabularnewline
Ours LPIPS interp & \textbf{997} & \textbf{0.200} \tabularnewline
\bottomrule
\end{tabular*}
\label{table:eval_mnist}
\end{table}

\subsection{Preliminary Study on Stacked MNIST}
\label{sec:preliminary_study_mnist}

In a real-world data distribution, the notion of modes is difficult to quantize. We instead start with Stacked MNIST~\cite{metz2016unrolled} where 1,000 discrete modes are unambiguously synthesized. This allows us to zoom in the challenge of mode collapse and facilitate a precise pre-validation.

We report the evaluation in Table~\ref{table:eval_mnist}. Our method narrows down the gap between experimental performance and the theoretical limit: It covers the most number of modes and achieves the closest mode distribution to the uniform distribution ground truth. This study validates the improved effectiveness of harmonizing IMLE with GAN, compared to the other GAN models or hybrid models, in terms of explicit mode/data coverage. This sheds the light and pre-qualifies to apply our method on more complicated real-world datasets.

\subsection{Empirical Study on Data and Model Biases}
\label{sec:bias_study}

As discussed in Section~\ref{sec:related_work}, data biases lead to biases in generative models. Even worse, a model without attention to minorities can exacerbate such biases against allocating adequate representation capacities to them. In this empirical study, we first show the existence of biases across CelebA attributes in terms of sample counts and sample variance, and then correlate them to the biased performance of the backbone StyleGAN2~\cite{karras2020analyzing}.

\begin{figure*}[!t]
\centering
\includegraphics[width=\textwidth]{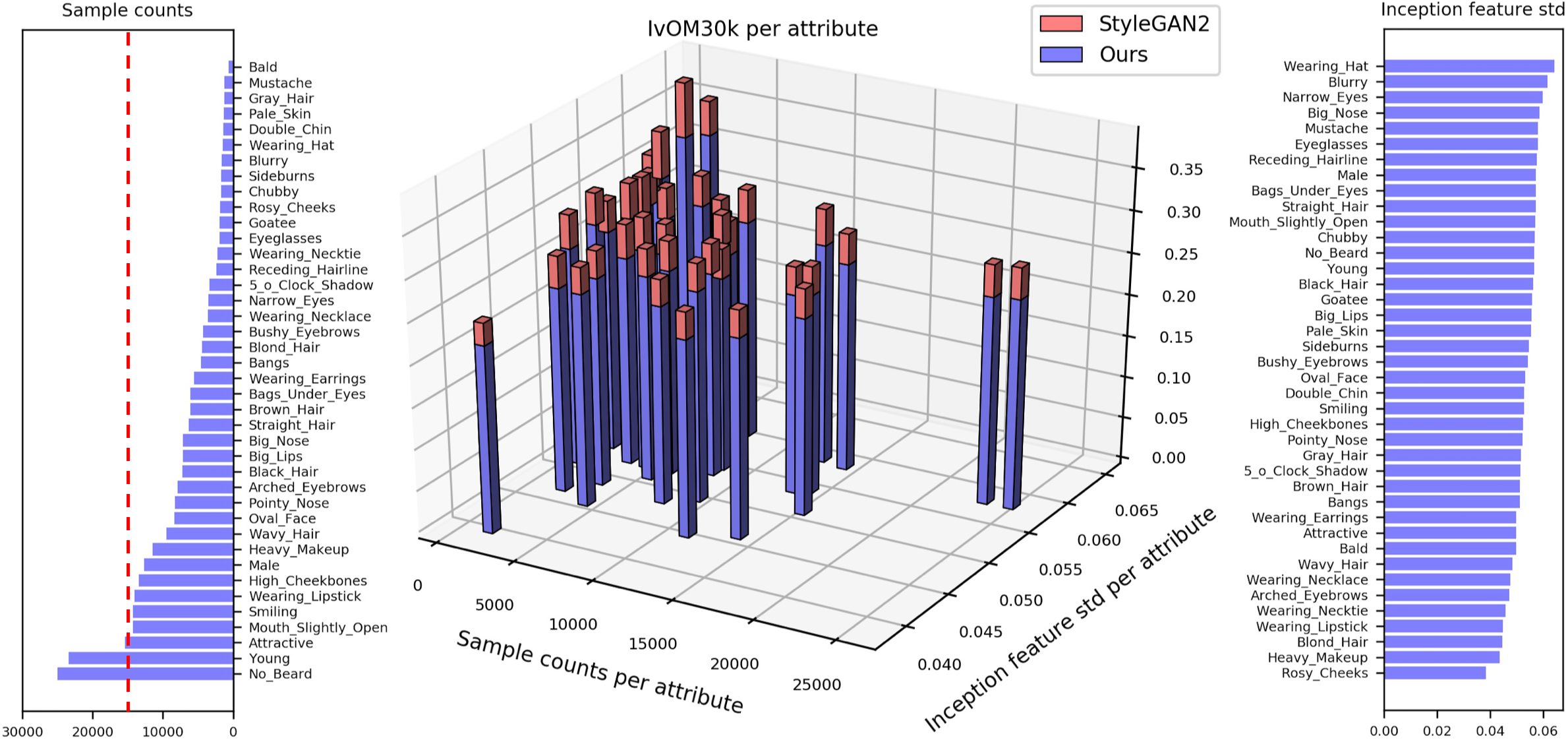}
\caption{Visualizations for data and model biases. Left: Sorted CelebA attribute histogram with a balance point marked by the red dashed line. Right: Sorted Inception feature variance per attribute. Middle: Per-attribute mean IvOM over 30,000 CelebA training samples for StyleGAN2 (red) and for our method (blue), where each bar corresponds to one attribute.}
\label{fig:counts_std_dists}
\end{figure*}

As shown in the left barplot of Figure~\ref{fig:counts_std_dists}, given the attribute histogram over 30,000 samples, 29 out of 40 binary attributes are more than 50\% biased from the balance point (15,000 out of 30,000 samples with a positive attribute annotation, shown as the red dashed line). On the other hand, in the right barplot of Figure~\ref{fig:counts_std_dists}, we calculate the standard deviation of Inception features~\cite{salimans2016improved} of samples within each attribute, and notice a wide range spanning from 0.038 to 0.062.

Too few samples or too large appearance variance in one attribute discourages generative coverage for that attribute, and thus results in biases. To quantify the per-attribute coverage, we measure the mean IvOM~\cite{metz2016unrolled} over positive training samples. A larger value indicates a worse coverage. In the middle barplot of Figure~\ref{fig:counts_std_dists}, we visualize the correlation between IvOM and the joint distribution of sample counts and sample variance. There is a clear gradient trend of IvOM when the samples of an attribute turn rarer and/or more diverse. To validate such a strong correlation, we first normalize the sample counts and sample variance across attributes by their means and standard deviations. Then we simply add them up as a joint variable vector, and calculate its Spearman's ranking correlation to the per-attribute IvOM. For StyleGAN2 (the red bar), the correlation coefficient of 0.75 indicates a strong correlation between data biases and model biases. This evidences the urgency to mitigate biases against the rare and diverse samples, in another word, to enhance the coverage over minority subgroups.

\begin{table}[!t]
\center
\caption{Comparisons on CelebA dataset. We indicate for each metric whether a higher ($\Uparrow$) or lower ($\Downarrow$) value is more desirable. The first part corresponds to the comparisons among different methods. For VAEGAN we report the results based on LPIPS distance metric. We report additional results based on the other three metrics in the supplementary material. We highlight the best performance in \textbf{bold} and the second best performance with \underline{underline}. We visualize the radar plots in Figure~\ref{fig:eval_radar_plot_main} for the comprehensive evaluation of each method over the validation set. The second part corresponds to our minority inclusion model variants in Section~\ref{sec:exp_minority_inclusion}.}
\begin{tabular*}{\textwidth}{@{\extracolsep{\fill}} lcccccccccc}
\toprule
& \multicolumn{2}{c}{FID30k} & \multicolumn{2}{c}{Precision30k} & \multicolumn{2}{c}{Recall30k} & \multicolumn{2}{c}{IvOM3k} & \multicolumn{2}{c}{IvOM3k std} \tabularnewline
& \multicolumn{2}{c}{$\Downarrow$} & \multicolumn{2}{c}{$\Uparrow$} & \multicolumn{2}{c}{$\Uparrow$} & \multicolumn{2}{c}{$\Downarrow$} & \multicolumn{2}{c}{$\Downarrow$} \tabularnewline
Method & Train & Val & Train & Val & Train & Val & Train & Val & Train & Val \tabularnewline
\midrule
StyleGAN2~\cite{karras2020analyzing} & \textbf{9.37} & \textbf{9.49} & 0.855 & 0.844 & 0.730 & 0.741 & 0.303 & 0.302 & 0.0268 & 0.0264 \tabularnewline
SNGAN~\cite{miyato2018spectral} & 13.32 & 13.24 & 0.792 & 0.787 & 0.631 & 0.616 & 0.325 & 0.322 & 0.0274 & 0.0261 \tabularnewline
Dist-GAN~\cite{tran2018dist} & 30.97 & 30.44 & 0.511 & 0.595 & 0.360 & 0.385 & 0.282 & 0.280 & 0.0220 & 0.0209 \tabularnewline
DSGAN~\cite{dsganICLR2019} & 14.29 & 14.00 & 0.868 & 0.862 & 0.679 & 0.724 & 0.301 & 0.300 & 0.0227 & 0.0220 \tabularnewline
PacGAN~\cite{lin2018pacgan} & 15.05 & 15.12 & 0.870 & \underline{0.869} & 0.726 & 0.758& 0.311 & 0.308 & 0.0256 & 0.0238 \tabularnewline
ALI~\cite{dumoulin2016adversarially} & \underline{10.09} & \underline{10.06} & 0.842 & 0.867 & 0.688 & 0.710 & 0.298 & 0.297 & 0.0240 & 0.0245 \tabularnewline
VAEGAN~\cite{larsen2015autoencoding} LPIPS & 24.10 & 23.47 & \underline{0.878} & 0.851 & 0.572 & 0.560 & 0.318 & 0.315 & 0.0284 & 0.0272 \tabularnewline
$\alpha$-GAN~\cite{rosca2017variational} & 12.65 & 12.53 & 0.803 & 0.810 & \underline{0.757} & \underline{0.763} & 0.267 & \underline{0.267} & 0.0208 & \underline{0.0192} \tabularnewline
VEEGAN~\cite{srivastava2017veegan} & 16.34 & 16.13 & 0.752 & 0.768 & 0.660 & 0.695 & \underline{0.260} & 0.269 & \textbf{0.0190} & \textbf{0.0181} \tabularnewline
Ours LPIPS interp & 11.56 & 11.28 & \textbf{0.927} & \textbf{0.941} & \textbf{0.849} & \textbf{0.848} & \textbf{0.255} & \textbf{0.262} & \underline{0.0193} & 0.0195  \tabularnewline
\midrule
Ours \textit{Eyeglasses} & 13.54 & 14.43 & 0.914 & 0.910 & 0.890 & 0.895 & 0.255 & 0.265 & 0.0249 & 0.0193 \tabularnewline
Ours \textit{Bald} & 13.34 & 13.46 & 0.903 & 0.895 & 0.886 & 0.892 & 0.268 & 0.272 & 0.0381 & 0.0227 \tabularnewline
Ours \textit{EN}\&\textit{HM} & 15.18 & 15.00 & 0.885 & 0.891 & 0.830 & 0.842 & 0.268 & 0.270 & 0.0318 & 0.0277 \tabularnewline
Ours \textit{BUE}\&\textit{HC}\&\textit{A} & 14.27 & 13.85 & 0.878 & 0.874 & 0.871 & 0.884 & 0.262 & 0.266 & 0.0300 & 0.0254 \tabularnewline
\bottomrule
\end{tabular*}
\label{table:eval_celeba_main}
\end{table}

\begin{figure*}[!t]
\centering
\includegraphics[width=\textwidth]{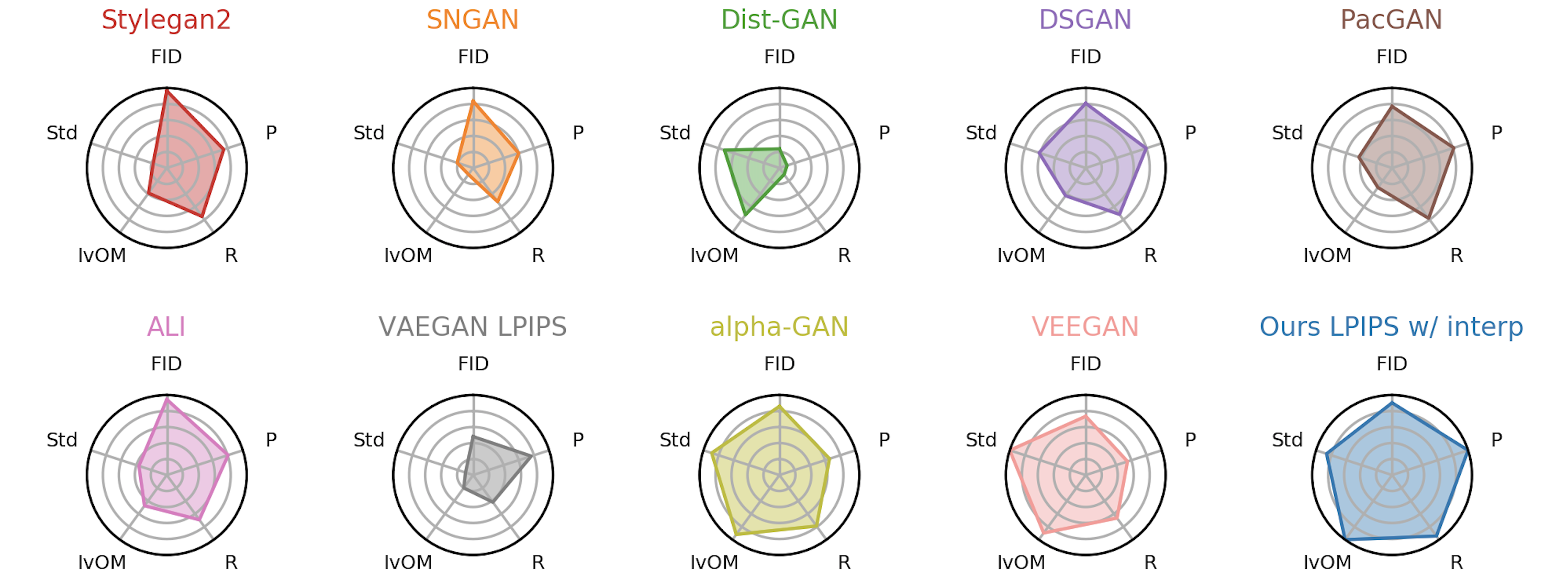}
\caption{Radar plots for the first part of Table~\ref{table:eval_celeba_main}.  ``P'' represents Precision, ``R'' represents Recall, and ``Std'' represents IvOM standard deviation. Values have been normalized to the unit range, and axes are inverted so that the higher value is always better.}
\label{fig:eval_radar_plot_main}
\end{figure*}

\begin{figure*}[!t]
\centering
\includegraphics[width=\textwidth]{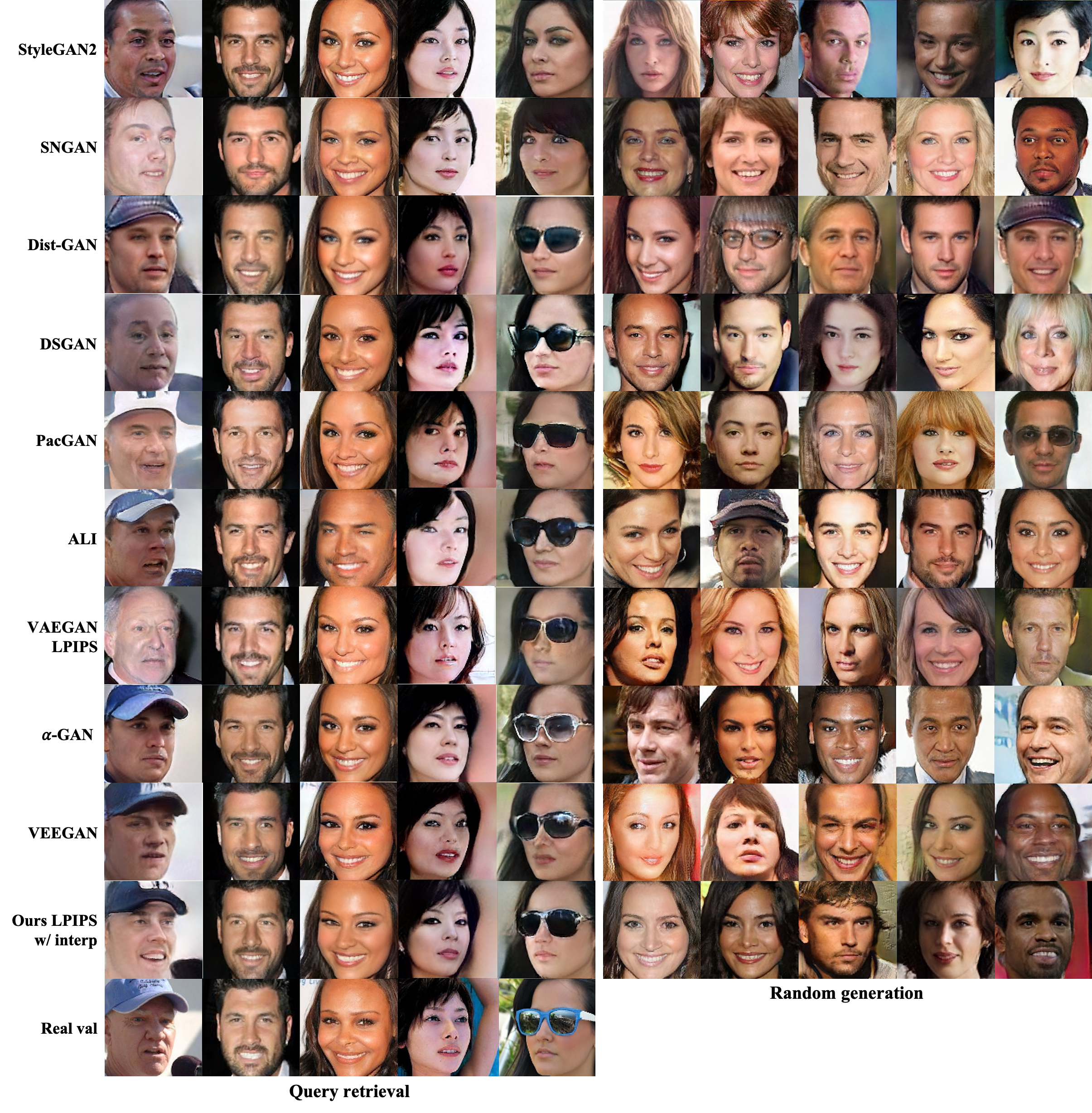}
\caption{Retrieval samples on the left (used for IvOM evaluation) and random generation samples on the right (used for FID, precision, and recall evaluation). The query images for retrieval in the top left row are real and unseen during training.}
\label{fig:rec_arb}
\end{figure*}

\subsection{Comparisons on CelebA}
\label{sec:comparisons_celeba}

In Section~\ref{sec:imle_gan} we propose two strategies to harmonize adversarial and reconstructive training: the deep distance metric and the interpolation-based augmentation. We compare four distance metrics and with/without augmentation in the supplementary material. We obtain: (1) LPIPS similarity shows near-top performance all around measures; and (2) interpolation-based augmentation consistently benefits all the measures in general for all the distance metrics. We therefore employ both into our full method.

To evaluate our data coverage performance in practice, we conduct comprehensive comparisons on CelebA~\cite{liu2015deep} against baseline methods. The first part of Table~\ref{table:eval_celeba_main} show our comparisons. Figure~\ref{fig:eval_radar_plot_main} assists interpret the table. We find:

(1) FID is not a gold standard to reflect the entire capability of a generative model, as it ranks differently from the other metrics.

(2) Compared to the original backbone StyleGAN2 which achieves the second-best FID, our full method (``Ours LPIPS interp'') trades slight FID deterioration for significant boosts in all the other metrics. This is meaningful because precision (FID) can be traded off at the expense of recall (Recall, IvOM) via the truncation trick used in~\cite{brock2018large,karras2020analyzing}, while the opposite direction is infeasible.

(3) Our full method outperforms all the existing state-of-the-art techniques in terms of Precision, Recall, and IvOM, where the latter two are the key evidence for effective data coverage. The last radar plot in Figure~\ref{fig:eval_radar_plot_main} shows our method achieves near-top measures all around with the most balanced performance.

(4) Our method also achieves the top-3 performance in the standard deviation of per-attribute IvOM, indicating an equalized capacity across the attribute spectrum. The blue bars in the middle barplot of Figure~\ref{fig:counts_std_dists} also visualize our method consistently outperforms StyleGAN2 (red bars) for all the attributes, in particular with more significant improvement for the minority subgroups.

(5) Figure~\ref{fig:rec_arb} shows qualitative comparisons in terms of query retrieval and uncurated random generation. StyleGAN2 suffers from mode collapse. For the collapsed modes, our method significantly improves the generation from non-existence of rare attributes to good quality (hat, sunglasses, etc.). Our method also demonstrates desirable generation fidelity and diversity.

(6) All the conclusions above generalize well to unseen data, as evidenced by the ``Val'' columns in Table~\ref{table:eval_celeba_main}.

\begin{table}[!t]
\center
\caption{Comparisons on CelebA minority subgroups, where the percentages show their portion w.r.t. the entire population. The metrics are measured on the corresponding subgroups only. We indicate for each metric whether a higher ($\Uparrow$) or lower ($\Downarrow$) value is more desirable. We highlight the best performance in \textbf{bold}.}
\begin{tabular*}{\textwidth}{@{\extracolsep{\fill}} llcccccc}
\toprule
& & \multicolumn{2}{c}{Precision1k} & \multicolumn{2}{c}{Recall1k} & \multicolumn{2}{c}{IvOM1k} \tabularnewline
& & \multicolumn{2}{c}{minority only} & \multicolumn{2}{c}{minority only} & \multicolumn{2}{c}{minority only} \tabularnewline
Arbitrary minority & & \multicolumn{2}{c}{$\Uparrow$} & \multicolumn{2}{c}{$\Uparrow$} & \multicolumn{2}{c}{$\Downarrow$} \tabularnewline
subgroup & Method & Train & Val & Train & Val & Train & Val \tabularnewline
\midrule
& StyleGAN2~\cite{karras2020analyzing} & 0.719 & 0.704 & 0.582 & 0.589 & 0.355 & 0.352 \tabularnewline
\textit{Eyeglasses} & Ours LPIPS interp & 0.843 & 0.845 & 0.740 & 0.708 & 0.309 & 0.308 \tabularnewline
(6\%)& Ours \textit{Eyeglasses} & \textbf{0.904} & \textbf{0.919} & \textbf{0.897} & \textbf{0.892} & \textbf{0.261} & \textbf{0.288} \tabularnewline
\midrule
& StyleGAN2~\cite{karras2020analyzing} & 0.707 & 0.750 & 0.461 & 0.424 & 0.301 & 0.305 \tabularnewline
\textit{Bald} & Ours LPIPS interp & 0.763 & \textbf{0.783} & 0.666 & 0.670 & 0.269 & \textbf{0.273} \tabularnewline
(2\%) & Ours \textit{Bald} & \textbf{0.779} & 0.718 & \textbf{0.842} & \textbf{0.810} & \textbf{0.189} & \textbf{0.273} \tabularnewline
\midrule
\textit{Narrow\_Eyes} & StyleGAN2~\cite{karras2020analyzing} & 0.719 & 0.701 & 0.543 & 0.577 & 0.272 & 0.274 \tabularnewline
\&\textit{Heavy\_Makeup} & Ours LPIPS interp & 0.794 & 0.760 & 0.632 & 0.621 & 0.246 & 0.248 \tabularnewline
(4\%) & Ours \textit{EN}\&\textit{HM} & \textbf{0.799} & \textbf{0.766} & \textbf{0.698} & \textbf{0.696} & \textbf{0.194} & \textbf{0.244} \tabularnewline
\midrule
\textit{Bags\_Under\_Eyes} & StyleGAN2~\cite{karras2020analyzing} & 0.838 & 0.804 & 0.736 & 0.725 & 0.263 & 0.268 \tabularnewline
\&\textit{High\_Cheekbones} & Ours LPIPS interp & 0.816 & 0.831 & 0.700 & 0.742 & 0.237 & 0.241 \tabularnewline
\&\textit{Attractive} (4\%) & Ours \textit{BUE}\&\textit{HC}\&\textit{A} & \textbf{0.889} & \textbf{0.883} & \textbf{0.813} & \textbf{0.809} & \textbf{0.191} & \textbf{0.237} \tabularnewline
\bottomrule
\end{tabular*}
\label{table:eval_minority}
\end{table}

\subsection{Extension to Minority Inclusion}
\label{sec:exp_minority_inclusion}

%Encouraged by the significant improvement in overall data coverage, 
% , so as to ensure explicit inclusion during generative modeling
We adapt our method for ensuring specific coverage over minority subgroups (Algorithm~\ref{alg:imle_gan}). Without introducing unconscious bias on the CelebA attributes, we arbitrarily specify four sets of attributes, the samples of which count for no more than 6\% of the population, and therefore, constitute four minority subgroups respectively. The attribute sets and their portions are listed in the first column of Table~\ref{table:eval_minority}.

\begin{figure*}[!t]
\centering
\includegraphics[width=\textwidth]{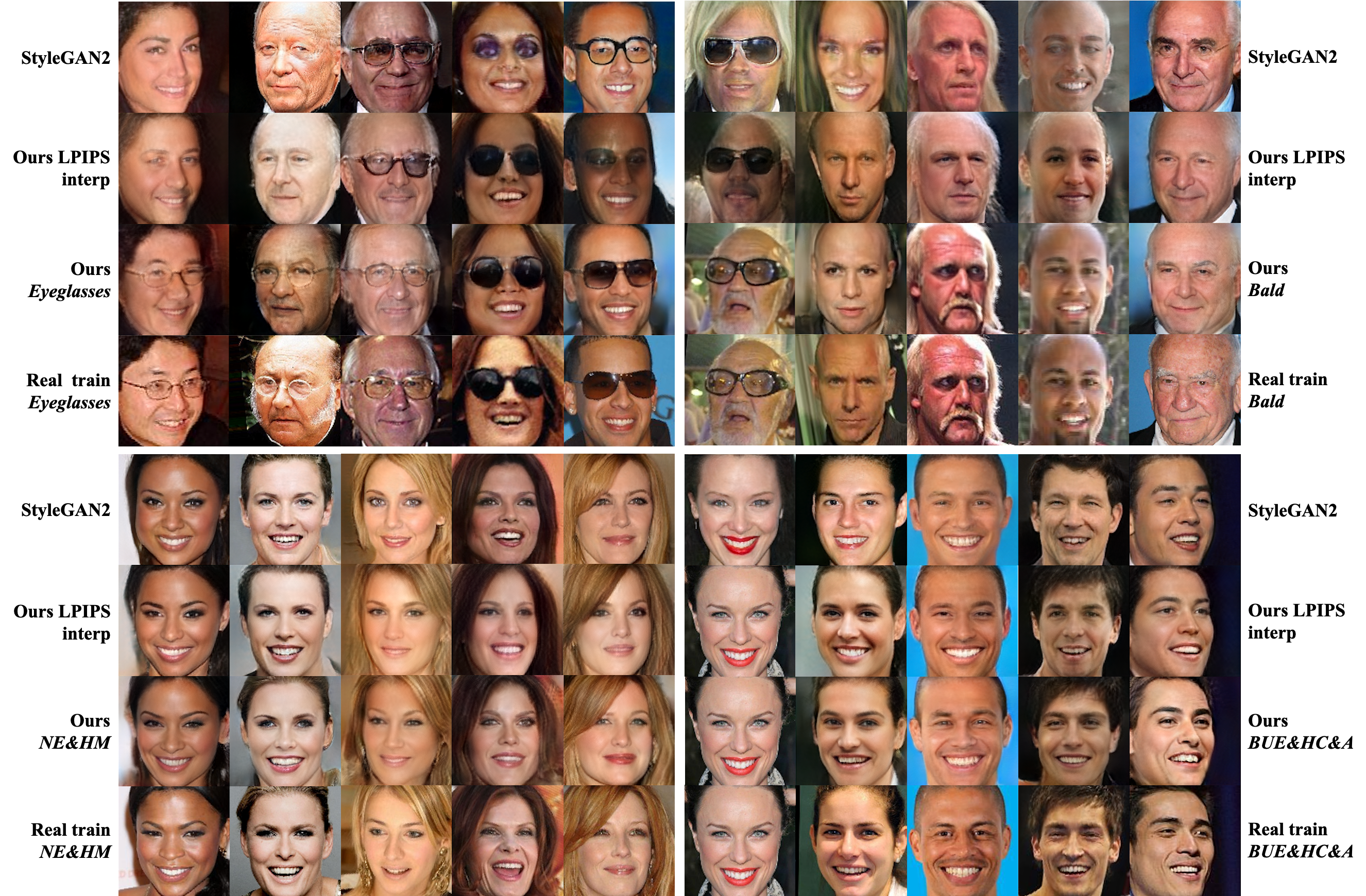}
\caption{Retrieval samples according to different minority subgroups. The query images for retrieval in the top row of each sub-figure are real from the training set.}
\label{fig:rec_minority}
\end{figure*}

To validate minority inclusion, we first compare our minority model variants over the corresponding minority subsets against the backbone StyleGAN2 and against our general full model. See Table~\ref{table:eval_minority} for the results. Our minority variants consistently outperform the two baselines over all the minority subgroups. In Figure~\ref{fig:rec_minority}, our method retrieves the minority attributes the most accurately, even for the subtle attributes like eye bags where StyleGAN2 fails. It validates better training data utilization of our minority models. Additional results are shown in the supplementary material.

To validate the overall performance beyond minority subgroups, we show at the bottom of Table~\ref{table:eval_celeba_main} the performance on the entire attribute spectrum. We conclude that the improvement of all our minority models comes at little or no compromise from their performance on the overall dataset. 

\section{Conclusion}

In this paper, we formalized the problem of minority inclusion as one of data coverage and improved data coverage using a novel paradigm that harmonizes adversarial training (GAN) with reconstructive generation (IMLE). Our method outperforms state-of-the-art methods in terms of Precision, Recall, and IvOM on CelebA, and the improvement generalizes well on unseen data. We further extended our method to ensure explicit inclusion for minority subgroups at little or no compromise on overall full-dataset performance. We believe this is an important step towards fairness in generative models, with the aim to reduce and ultimately prevent discrimination due to model and data biases. %We demonstrate this is an effective step towards more inclusive generative modeling.

\section{Acknowledgement.}

This project was partially funded by DARPA MediFor program under cooperative agreement FA87501620191 and by ONR MURI (N00014-14-1-0671). We thank Tero Karras and Michal Luk\'{a}\v{c} for sharing code. We also thank Richard Zhang and Dingfan Chen for constructive advice in general.

\bibliographystyle{splncs04}
\bibliography{main}

\clearpage
\renewcommand\thesubsection{\Alph{subsection}}

\section{Supplementary material}

\subsection{Implementation Details}

\noindent\textbf{IMLE-GAN.} We train each model using Adam optimizer~\cite{kingma2015adam} for $K = 300$ epochs. We use no exponential decay rate ($\beta_1=0.0$) for the first moment estimates, and use the exponential decay rate $\beta_2=0.99$ for the second moment estimates. The learning rate $\eta = 0.002$, the same as in StyleGAN2~\cite{karras2020analyzing}. We update the matching of latent vectors to data points every $S = 20$ epochs. The size of the pool of latent vector candidates is 10 times of the size of the dataset or the minority group depending on the application. Perturbation variance is $\sigma^2 = 0.05^2$. The weight of the reconstruction loss varies according to the choice of metric, such that the magnitude of the reconstruction loss is about equal to that of the adversarial loss. For $\ell_2$ distance in pixel space, $\lambda = 36$. In the discriminator feature space~\cite{larsen2015autoencoding} $\lambda = 9.6\times 10^6$. In the Inception feature space~\cite{heusel2017gans} $\lambda = 10$. For LPIPS~\cite{zhang2018unreasonable} $\lambda = 2.5$. Consistently, the weight of the interpolation loss is always set to $\beta = 0.4\lambda$.

We train all our models on 3 NVIDIA V100 Tensor Core GPUs with 16GB memory each. Based on the memory available and the training performance, we set the batch size at 32 for the 240,000 32$\times$32$\times$3 Stacked MNIST images~\cite{metz2016unrolled}, and the training lasts for 1.7 days. We set the batch size at 16 for the 30,000 128$\times$128$\times$3 CelebA images~\cite{liu2015deep}, and the training lasts for 2.4 days.

\noindent\textbf{Baseline methods.} For fair comparisons, all the baseline methods are re-implemented using the same StyleGAN2 backbone and training strategies. For ALI~\cite{dumoulin2016adversarially}, VAEGAN~\cite{larsen2015autoencoding}, $\alpha$-GAN~\cite{rosca2017variational}, and VEEGAN~\cite{srivastava2017veegan} where an encoder is involved, we adapt the discriminator architecture for the encoder. For Dist-GAN~\cite{tran2018dist}, we measure image distance by LPIPS~\cite{zhang2018unreasonable} and tune the weight of the distance constraint term such that its value is about 1/4 of the adversarial loss. For DSGAN~\cite{dsganICLR2019}, we tune the weight of the diversity regularization term such that its value is about 1/4 of the adversarial loss. For PacGAN~\cite{lin2018pacgan}, we set the pack size to 8. For VAEGAN~\cite{larsen2015autoencoding}, we tune the weights of the data reconstruction term and the prior term such that the former is about equal to the adversarial loss and the latter is about 1/4 of that. For $\alpha$-GAN~\cite{rosca2017variational}, we use LPIPS distance~\cite{zhang2018unreasonable} to reconstruct images and tune the weight of the reconstruction term such that its value is about 1/4 of the adversarial loss. For VEEGAN~\cite{srivastava2017veegan}, we tune the weight of the latent reconstruction term such that its value is about equal to the adversarial loss. For SNGAN~\cite{miyato2018spectral} and ALI~\cite{dumoulin2016adversarially}, there is no additional hyperparameter.

\noindent\textbf{Evaluation.} For Precision and Recall~\cite{sajjadi2018assessing} measurement, we use the default setting from their official code repository. In particular, the features are extracted from the \textit{Pool3} layer of a pre-trained Inception network~\cite{heusel2017gans}. The number of clusters for $k$-means is set to 20. We launched for 10 independent runs and report the average for Precision and Recall. For IvOM~\cite{metz2016unrolled} measurement, the retrieval is implemented as an optimization w.r.t. the latent vector, such that a learned generator approximates its generation towards the query image. The retrieval error is then calculated as the difference between the optimal generated image and the query image. The optimization objective and the error are measured using the deep similarity metric LPIPS~\cite{zhang2018unreasonable}. Given each query image and a learned generative model, we optimized the latent vector via Adam~\cite{kingma2015adam} for 400 steps. The learning rate setting strategy is the same as in StyleGAN2: the maximum learning rate is 0.1, and it is ramped up from zero linearly during the first 20 steps and ramped down to zero using a cosine schedule during the last 100 steps.

\subsection{Effectiveness of Harmonization}

In Section~3.3 in the main paper, we propose two strategies to harmonize adversarial and reconstructive training: the deep distance metric and the interpolation-based augmentation. We compare four distance metrics and with/without augmentation in the third part of Table~\ref{table:eval_celeba_supp}. For distance metrics, the pixel space (the vanilla version) achieves the desirable Recall and the Inception space achieves the desirable FID, but they contain obvious shortcomings in the other measures. In contrast, the LPIPS similarity shows near-top measures all around with the most balanced performance, which is employed in our full method. For augmentation, it consistently benefits all the measures in general for all the distance metrics, which is also employed. In summary, harmonizing GAN and IMLE is a non-trivial challenge. Our two strategies achieve the best of the two worlds by significantly improving the overall performance (including data coverage) from the vanilla version.

For completeness, in Table~\ref{table:eval_celeba_supp} second part we also compare to VAEGAN which is alternatively incorporated with different distance metrics. Although LPIPS metric boosts our method the most, we find Inception space boosts VAEGAN the most. But it is still not as advantageous as our performance in general, especially for Recall and IvOM which corresponds to data coverage.

\begin{figure*}[!t]
\centering
\includegraphics[width=\textwidth]{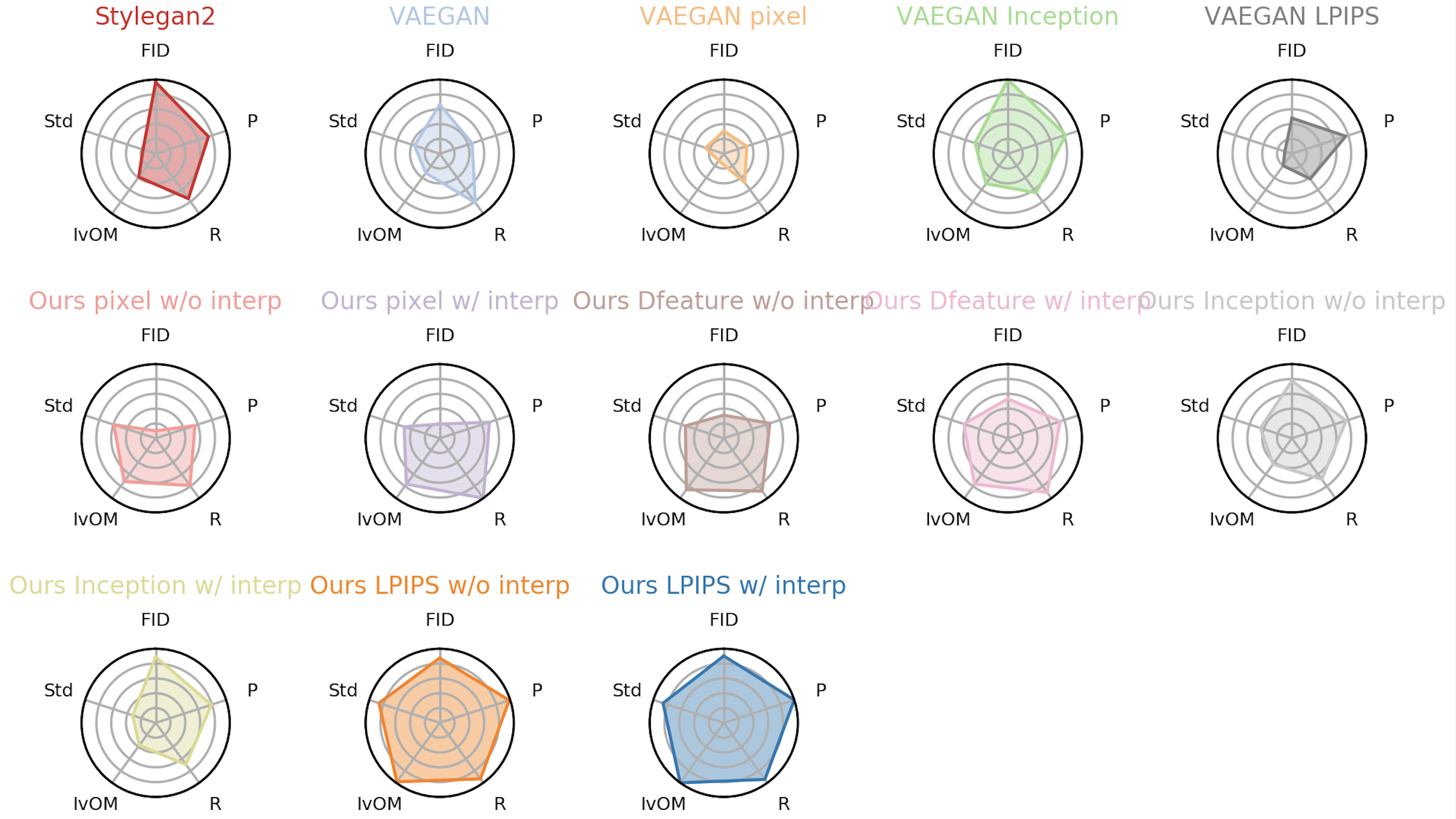}
\caption{Radar plots for Table~\ref{table:eval_celeba_supp}.  ``P'' represents Precision, ``R'' represents Recall, and ``Std'' represents IvOM standard deviation. Values have been normalized to the unit range, and axes are inverted so that the higher value is always better.}
\label{fig:eval_radar_plot_supp}
\end{figure*}

The radar plots in Figure~\ref{fig:eval_radar_plot_supp} assist interpret Table~\ref{table:eval_celeba_supp}.

\begin{table}[!t]
\center
\caption{Comparisons on CelebA dataset. We indicate for each metric whether a higher ($\Uparrow$) or lower ($\Downarrow$) value is more desirable. We highlight the best performance in \textbf{bold} and the second best performance with \underline{underline}. We visualize the radar plots in Figure~\ref{fig:eval_radar_plot_supp} for the comprehensive evaluation of each method over the validation set.}
\begin{tabular*}{\textwidth}{@{\extracolsep{\fill}} lcccccccccc}
\toprule
& \multicolumn{2}{c}{FID30k} & \multicolumn{2}{c}{Precision30k} & \multicolumn{2}{c}{Recall30k} & \multicolumn{2}{c}{IvOM3k} & \multicolumn{2}{c}{IvOM3k std} \tabularnewline
& \multicolumn{2}{c}{$\Downarrow$} & \multicolumn{2}{c}{$\Uparrow$} & \multicolumn{2}{c}{$\Uparrow$} & \multicolumn{2}{c}{$\Downarrow$} & \multicolumn{2}{c}{$\Downarrow$} \tabularnewline
Method & Train & Val & Train & Val & Train & Val & Train & Val & Train & Val \tabularnewline
\midrule
StyleGAN2~\cite{karras2020analyzing} & \underline{9.37} & \underline{9.49} & 0.855 & 0.844 & 0.730 & 0.741 & 0.303 & 0.302 & 0.0268 & 0.0264 \tabularnewline
\midrule
VAEGAN~\cite{larsen2015autoencoding} & 18.26 & 18.14 & 0.738 & 0.733 & 0.782 & 0.779 & 0.310 & 0.307 & 0.0264 & 0.0246 \tabularnewline
VAEGAN pixel & 28.89 & 28.49 & 0.689 & 0.683 & 0.573 & 0.594 & 0.323 & 0.320 & 0.0259 & 0.0256 \tabularnewline
VAEGAN Inception & \textbf{8.35} & \textbf{8.47} & 0.875 & 0.872 & 0.687 & 0.687 & 0.298 & 0.295 & 0.0248 & 0.0235 \tabularnewline
VAEGAN LPIPS & 24.10 & 23.47 & 0.878 & 0.851 & 0.572 & 0.560 & 0.318 & 0.315 & 0.0284 & 0.0272 \tabularnewline
\midrule
Ours pixel & 34.94 & 34.46 & 0.774 & 0.771 & 0.751 & 0.763 & 0.272 & 0.280 & 0.0199 & 0.0222 \tabularnewline
Ours pixel interp & 32.54 & 31.82 & 0.828 & 0.828 & \textbf{0.882} & \textbf{0.879} & 0.265 & 0.277 & 0.0207 & 0.0231 \tabularnewline
Ours Dfeature~\cite{larsen2015autoencoding} & 28.85 & 28.34 & 0.793 & 0.808 & 0.811 & 0.814 & \textbf{0.255} & 0.271 & \textbf{0.0188} & 0.0227 \tabularnewline
Ours Dfeature interp & 22.38 & 21.92 & 0.849 & 0.842 & 0.806 & 0.826 & 0.263 & 0.277 & \underline{0.0189} & 0.0219 \tabularnewline
Ours Inception~\cite{salimans2016improved} & 14.86 & 14.95 & 0.859 & 0.853 & 0.675 & 0.706 & 0.294 & 0.299 & 0.0232 & 0.0237 \tabularnewline
Ours Inception interp & 11.62 & 11.61 & 0.843 & 0.861 & 0.704 & 0.712 & 0.301 & 0.303 & 0.0234 & 0.0249 \tabularnewline
Ours LPIPS~\cite{zhang2018unreasonable} & 12.30 & 12.10 & \underline{0.916} & \underline{0.936} & 0.835 & 0.843 & 0.256 & \underline{0.263} & 0.0194 & \textbf{0.0195} \tabularnewline
Ours LPIPS interp & 11.56 & 11.28 & \textbf{0.927} & \textbf{0.941} & \underline{0.849} & \underline{0.848} & \textbf{0.255} & \textbf{0.262} & 0.0193 & \textbf{0.0195}  \tabularnewline
\bottomrule
\end{tabular*}
\label{table:eval_celeba_supp}
\end{table}

\subsection{Additional Results on Minority Inclusion}

In order to dynamically demonstrate the effectiveness of our minority inclusion models, we are attaching four videos at \href{https://github.com/ningyu1991/InclusiveGAN.git}{GitHub}. The videos show the results of interpolating in the latent space from one arbitrary image to another image with specific attribute(s). In this way we show our minority inclusion model variants perform comparably to the other models for majority groups, and outperform the others for minority groups.

\begin{figure*}[!t]
\centering
\subfigure[Minority: \textit{Eyeglasses}]{\includegraphics[width=\textwidth]{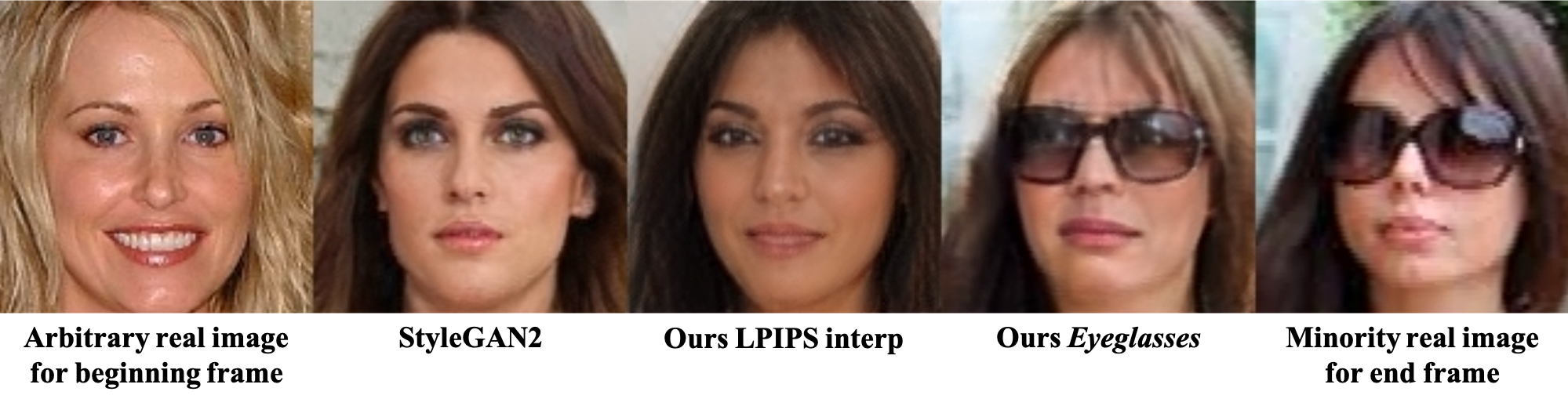}}
\subfigure[Minority: \textit{Bald}]{\includegraphics[width=\textwidth]{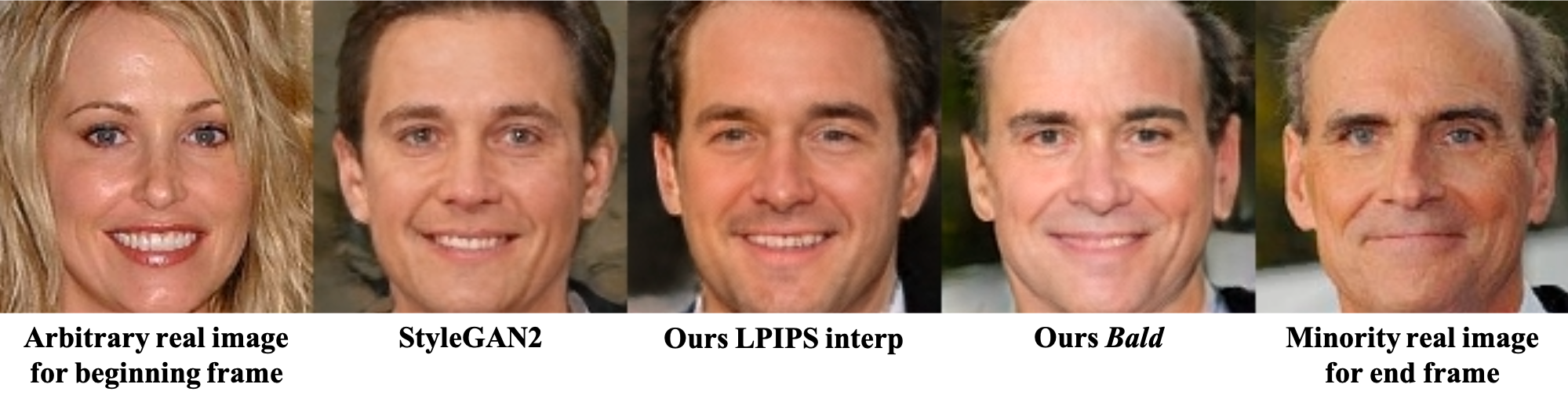}}
\subfigure[Minority: \textit{Narrow\_Eyes}\&\textit{Heavy\_Makeup}]{\includegraphics[width=\textwidth]{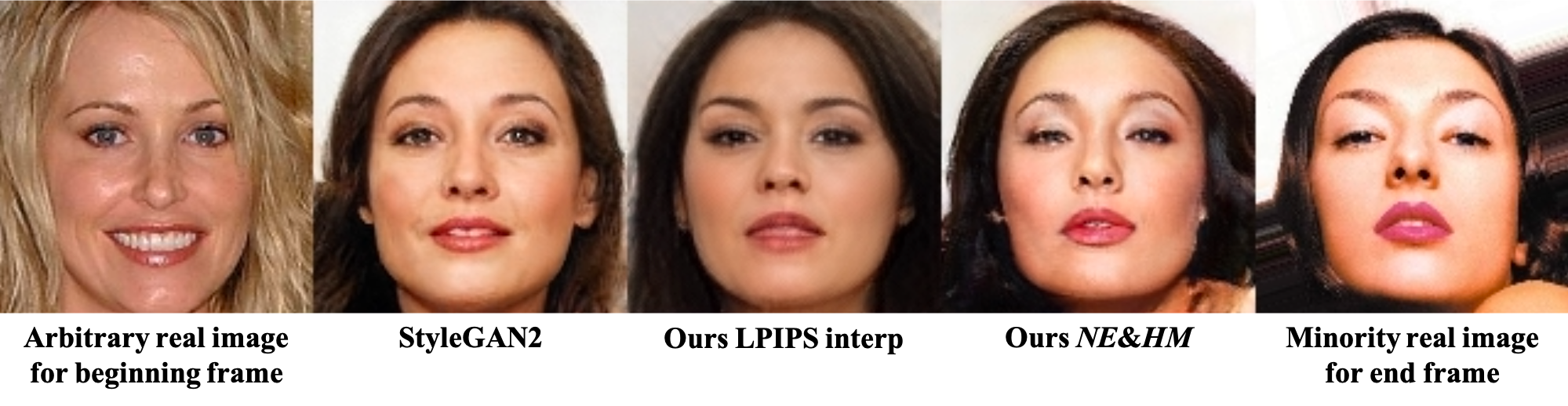}}
\subfigure[Minority: \textit{Bags\_Under\_Eyes}\&\textit{High\_Cheekbone}\&\textit{Attractive}]{\includegraphics[width=\textwidth]{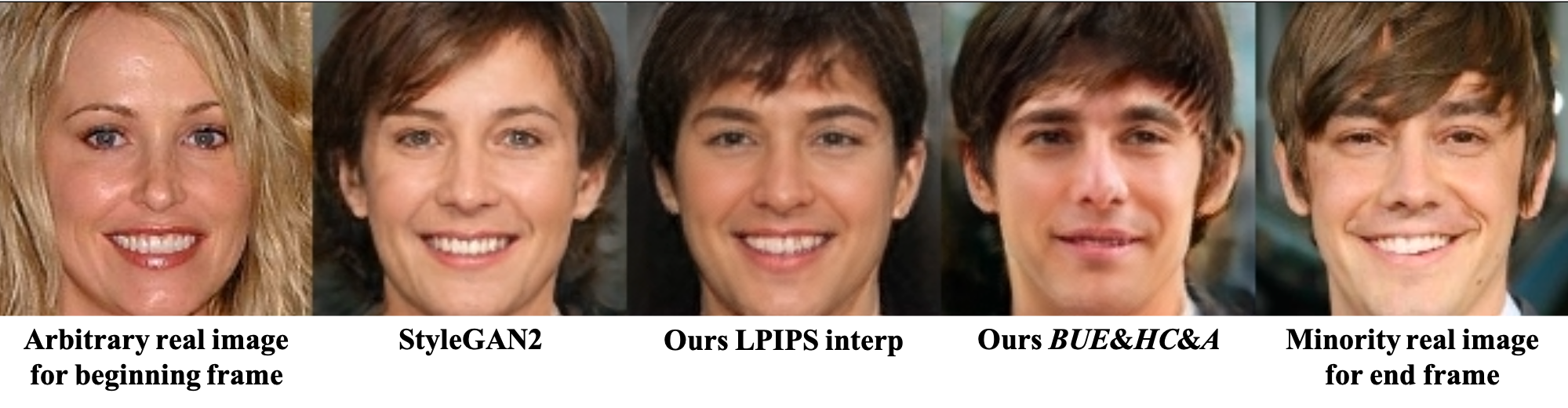}}
\caption{The last frame of each video in the attachment and also at \href{https://github.com/ningyu1991/InclusiveGAN.git}{GitHub}. Each of the middle three columns denotes a generated image from a learned model, the latent vector of which is projected from the image in the rightmost column (a real image from one minority subgroup).}
\label{fig:video_last_frame}
\end{figure*}

In each video, the leftmost column is an arbitrary real image and the rightmost column is an arbitrary real image with specific attribute(s) of interest. For each generative model, we project the image in the leftmost column onto its latent space (i.e.: we find the latent vector that results in a generated image that is most perceptually similar to the image according to LPIPS~\cite{zhang2018unreasonable}), and then interpolate starting from this latent vector. We do the same for the image in the rightmost column and use the resulting latent vector as the target for interpolation. The sub-videos in the middle three columns are the images produced by three methods: StyleGAN2~\cite{karras2020analyzing}, our general IMLE-GAN model described in Section~3.3 and 4.4 in the main paper~(``Ours LPIPS interp''), and our IMLE-GAN model with specific minority inclusion described in Section~3.4 and 4.5 in the main paper~(Ours \textit{attributeA}\&\textit{attributeB}). The four videos correspond to the four arbitrarily selected attributes or attribute combinations used in Section~4.5 in the main paper: \textit{Eyeglasses}, \textit{Bald}, \textit{Narrow\_Eyes}\&\textit{Heavy\_Makeup}~(\textit{NE}\&\textit{HM}), and \textit{Bags\_Under\_Eyes}\&\textit{High\_Cheekbone}\&\textit{Attractive}~(\textit{BUE}\&\textit{HC}\&\textit{A}). For convenience, we show the last frame of each video in Figure~\ref{fig:video_last_frame}, where each generated image is the projection of the rightmost image (a real image from the minority group) onto the space of images learned by each generative model.

We note from the qualitative comparisons that incorporating minority inclusion in the training objective ensures coverage of the specified minority group, with little or no compromise from their performance on the majority. For example, in each video, at the beginning the three models are comparably representative for the arbitrary real image from the majority group (the leftmost column). As the latent vector transitions towards the corresponding minority region (the rightmost column), the attribute appearances of the minority group are not reconstructed accurately by the two models without an explicit focus on minority attributes (the second and third columns from the left). On the contrary, our minority inclusion model (the second column from the right) effectively represents the desired minority attributes, e.g., sunglasses, narrow eye shapes, or eye bags.

\end{document}